\documentclass{article}

\PassOptionsToPackage{numbers, sort&compress}{natbib}

\usepackage[final]{neurips_2022}

\usepackage[utf8]{inputenc} 
\usepackage[T1]{fontenc}    
\usepackage{hyperref}       
\usepackage{url}            
\usepackage{booktabs}       
\usepackage{amsfonts}       
\usepackage{mathtools}      
\usepackage{amsmath}        
\usepackage{booktabs}       
\usepackage{multirow}       
\usepackage{nicefrac}       
\usepackage{microtype}      
\usepackage{xcolor}         
\usepackage{graphicx}
\usepackage{caption}
\usepackage{subcaption}
\usepackage[super]{nth}
\usepackage{wrapfig}

\graphicspath{figures/}

\title{Explainability Via Causal Self-Talk}

\author{%
  Nicholas A. Roy\thanks{Equal contribution} \\
  DeepMind\\
  \texttt{nroy@deepmind.com} 
  \And
  Junkyung Kim\footnotemark[1]\\
  DeepMind\\
  \texttt{junkyung@deepmind.com} 
  \And
  Neil Rabinowitz\footnotemark[1]\ \ \thanks{Corresponding author}\\
  DeepMind\\
  \texttt{ncr@deepmind.com} \\
}

\begin{document}

\maketitle


\begin{abstract}

Explaining the behavior of AI systems is an important problem that, in practice, is generally avoided.
While the XAI community has been developing an abundance of techniques, most incur a set of costs that the wider deep learning community has been unwilling to pay in most situations.
We take a pragmatic view of the issue, and define a set of desiderata that capture both the ambitions of XAI and the practical constraints of deep learning.
We describe an effective way to satisfy all the desiderata: train the AI system to build a causal model of itself.
We develop an instance of this solution for Deep RL agents: Causal Self-Talk.
CST operates by training the agent to communicate with itself across time.
We implement this method in a simulated 3D environment, and show how it enables agents to generate faithful and semantically-meaningful explanations of their own behavior.
Beyond explanations, we also demonstrate that these learned models provide new ways of building semantic control interfaces to AI systems.

\end{abstract}

\section{Introduction}

As modern machine learning systems become more powerful and embedded in our lives, the need to have these systems explain their behavior becomes increasingly urgent. Despite incredible performance in a variety of domains, almost all systems are completely unable to provide a satisfying answer to the simple question, "Why did you do that?".

While there have been innovations in particular explainability methods \citep[e.g.][]{Bach2015-lz, Mordvintsev2015-bw, Topin2019-nz, Mott2019-kg}, and some attempts to apply techniques to domains that demand it \citep[e.g.][]{Fauw2018-rm, Tenney2020-zb}, the vast majority of deep learning applications do not engage with these techniques. This is in part because there are few general methods, and most incur costs such as reduced performance and scalability; 
compounding this, existing techniques often fail to deliver a useful or credible account of the systems' behavior. 

We believe that the reason for this is that explanation systems tend to fail to satisfy at least one of five key desiderata discussed in detail below, namely: groundedness, flexibility, minimally-interfering, scalability, and faithfulness.
Two particular failure modes are most common: the addition of an explanation system limits generality or hurts performance (e.g.\ when it requires explicitly structuring the base system); or the explanations it produces cannot be trusted to be accurate (e.g.\ with many external or post-hoc methods).
An ideal explanation system would avoid both these pitfalls.

In this paper, we propose one such solution. We take a pragmatic viewpoint that providing explanations of a system's behavior is, in essence, a task of building a \textit{model} of that system. 
We argue that the AI system itself is well-positioned to supply a model that fulfills all the aforementioned desiderata.
We thus train the base system to supply a "self-model" alongside its main representations.
Importantly, the base system and self-model are subject to different constraints: we want the base system to be primarily optimized for performance, while we want the self-model to take on a faithful, semantically-interpretable form. The different form allows the self-model to deliver supplementary utility that the base system does not, e.g.\ providing an interface to external users, and enabling them to understand, predict, and control the base system in convenient ways.

Our contributions are as follows. In Section~\ref{section-desiderata}, we articulate five key desiderata for explanation systems. However, there is some tension between these. 
In Section~\ref{section-historical-solutions}, we describe how common historical XAI techniques resolve these trade-offs by making extreme choices between the desiderata. 
Section~\ref{section-solution-concept} marks a transition from discussing the desiderata in theory to exploring them in practice, as we demonstrate that a balance can be found between the five desiderata by training AI systems to build self-models. In particular, we introduce a solution concept for Deep RL agents: Casual Self-Talk (CST), in which an agent must learn to use its own outputs as inputs. In Sections~\ref{section-experiments}-\ref{section-results} we show that CST allows model-free RL agents in an embodied 3D virtual environment to faithfully explain their behavior in terms of semantically-accessible beliefs about the state of the world.
\section{Desiderata for explanatory models}
\label{section-desiderata}

Our goal is to build an explanatory model of an AI system, i.e.\ a mechanism that translates between its internal representations and/or computations, and some other representational form. We set out five key desiderata for such a model. We flesh these out in more detail in Section~\ref{section-historical-solutions}.

\begin{changemargin}{0.5cm}{0.5cm}
\textbf{Grounded.  } For the model to be useful to external users, it should deploy a representation language that is semantically-grounded. This may require additional data to train.
\end{changemargin}

\begin{changemargin}{0.5cm}{0.5cm}
\textbf{Flexible.  } The method should be able to produce models of many different representational forms (e.g. classes, graphs, strings). Ideally it should also be sufficiently general to be agnostic to the base system's input/output modalities, and possibly even its overall architecture.
\end{changemargin}

\begin{changemargin}{0.5cm}{0.5cm}
\textbf{Minimally-interfering.  } Training and inference on the model should have as little impact as possible on the training and inference of the base system. Solutions which are disruptive to performance are highly unlikely to be adopted in practice.
\end{changemargin}

\begin{changemargin}{0.5cm}{0.5cm}
\textbf{Scalable.  } Once a solution is found, it is straightforward to apply it to any size network.
\end{changemargin}

\begin{changemargin}{0.5cm}{0.5cm}
\textbf{Faithful.  } The model should provide accurate descriptions of the underlying system. The gold standard is a \textit{causal model} of the agent, which facilitates validation of its accuracy through interventions~\citep{Jain2019-wt}. This is possibly the most challenging to achieve, and is discussed further in the specific context of decoders in Section~\ref{section-historical-solutions}.
\end{changemargin}

A number of past works have also articulated desirable properties for explanations \citep[e.g.][]{Lipton2018-at, Guidotti2018-cf, Jain2019-wt, Wiegreffe2019-oa, Miller2019-la, Jacovi2020-hd, Jacovi2021-nl}.\footnote{Note that some terms, such as `faithfulness', do not have standard definitions across this literature \citep{Jacovi2020-hd}.} Our desiderata focus less on the qualities which constitute a good explanation per se, but instead on what is desired of the method itself that generates the explanation.

\section{Common historical solutions}
\label{subsection-non-solutions}
\label{section-historical-solutions}

We examine whether various existing explanation techniques satisfy each of these desiderata.

\paragraph{Do nothing.}
In standard Deep Learning (DL), we take the base system's learned internal representations at face value. Using the base system as a model of itself clearly has benefits of being minimally-interfering (no changes are made), scalable (as much as DL itself), and faithful (since the internal representations are causal to the base system's output). However, the approach entirely sacrifices grounding and flexibility.

\paragraph{Attribution methods.}
A set of popular techniques attempts to model a neural network's decision-making at the input level, by estimating contributions to its output from input features \citep[e.g.][]{Mott2019-kg, Goyal2019-ev, Cho2020-sd}, or training examples \citep[e.g.][]{Yeh2018-xx, Bien2011-xx, Kim2014-xx, Kim2016-xx, Koh2017-xx, Anirudh2017-xx},
or by finding optimal inputs for a given output \citep[e.g.][]{Zeiler2014-wy, Mordvintsev2015-bw}. Such approaches benefit from being minimally-interfering (they are entirely post-hoc) and scalable (they can be automated).
However, some methods have been empirically shown to be unfaithful \citep{Adebayo2018-mg, Atrey2019-ma}, while methods which explicitly aim for faithfulness such as LIME \citep{Biecek2021-va} and Anchors \citep{Ribeiro2018-xx} are only local in scope, and are expensive to scale. Most importantly, all such methods sacrifice flexibility: one is constrained to produce explanations of a very particular form, e.g.\ expressed in terms of the input modality, whose interpretation may be highly subjective. 

\paragraph{Explicitly structuring networks.} A sub-field of deep learning has emerged wherein the intermediate computations are constrained to particular forms, and/or are grounded by external data \citep[e.g.][]{Zambaldi2018-rl, Battaglia2018-xw, Burgess2019-aq, Andreas2016-ao, Shu2017-cb, Fauw2018-rm, Koh2020-mp, Geiger2021-cd, Rudin2019-ua}. 
As explanations of behavior, structured intermediates are attractive as they are both grounded and faithful by design. However, this is achieved at the expense of minimal-interference, as it requires fundamentally altering the base system to explicitly depend on this structure. It is also limited in flexibility and scalability, as the technique can also only be applied when sufficient domain knowledge or data is available, and only in the form dictated by these prior constraints. These issues are exacerbated by the fact that the representations required to optimize performance objectives are often at odds with those that would subserve explainability.

\paragraph{Fine-grained mechanistic interpretability.} One common approach is to allow a base system to train and run without interference, and to post-hoc dissect its representations or computations, often using techniques borrowed or inspired from neuroscience \citep[e.g.][]{Cammarata2020-my, Goh2021-fw,  Elhage2021-gn, Olsson2022-wo}. Such approaches champion the minimal-interference desideratum. In the best case, they promise to uncover faithful descriptions of the actual causal processes operating within the base system. However, this comes at a huge cost to scalability. Many human hours are typically expended in reverse engineering, often yielding results that are limited to a handful of neurons or circuits seen in a single network. Moreover, by making strong commitments to respect the internal representations of the base system, it is very difficult to ground the models that arise, let alone to flexibly choose the form they take.

\paragraph{Decoders.} Decoders can be viewed as trained tools to map from the internal private language of a base system to an interpretable representation space \citep[e.g.][]{Hendricks2016-pa, Bau2017-lx, Kim2017-dr, Zhou2018-gs, Alvarez-Melis2018-xo, Park2018-ty, Hendricks2018-af, McGrath2021-qx, Bonnen2021-zy, Hendricks2021-af}. This approach has many attractive qualities: it produces semantically-grounded representations of the base system's internal private language; is flexible in the choice of modality; scales well; and is minimally-interfering, insofar as decoders can be separated from the inference path of the main computational graph and gradients can be stopped from propagating to the base system if desired (though are often not to assist representation learning, \citep[e.g.][]{Lample2017-em, Jaderberg2017-pw, Mirowski2017-ay, Akata2018-al, Kartal2019-mj}). This is all achieved, however, at the expense of faithfulness, as decoded outputs are off the causal path of the base system \citep{Koh2020-mp}. Decoders just report whether information is \textit{present} in the source representation, not whether that information is actually being \textit{used} for behavior \citep{Belinkov2019-in, Ravichander2020-rc, Elazar2021-zx}. Similar faithfulness issues arise with post-hoc behavioral analyses \citep[e.g.][]{Hayes2017-gk, Juozapaitis2019-ag, Topin2019-nz, Yau2020-bz, Madumal2020-ta, Deletang2021-xy}.

\section{Solution concept}
\label{section-solution-concept}

The challenges faced by previous attempts at explainability show that the tensions between the desiderata are considerable.
It is nonetheless possible to reason a way forward.

\paragraph{Motivation.} First, we want a description of the base system that is both grounded and flexible. The internals of the base system are neither. While we could explicitly structure the base system to change this, this would violate the minimal interference desideratum. Thus we need \textit{a model} of the base system. 
Second, the solution must be scalable. This means we cannot rely on human labor to map from the base system's representations to the model's representations. The only scalable approach is to \textit{learn the model}.
Third, the solution must be faithful. 
Our gold standard of faithfulness requires the ability to validate the model by intervening on it.
To do this, one must be able to map in the opposite direction: from interventions on the model onto interventions on the base system.
Because both the base system and model are learned, this mapping between them must also be learned.
The most direct way of learning this mapping from model to base system is to \textit{feed the model's outputs as inputs to the base system}.

A solution to these problems is thus a ``self-model'' which is intertwined with the base system.
We develop an instance of this solution in a particular setting:
a Deep RL agent, acting in an episodic POMDP, which builds a model of itself in terms of a set of semantically-grounded beliefs about the state of the world. We consider how to expand the scope of this technique in Section~\ref{section-discussion}.

\paragraph{Agent construction.} We follow the standard construction of a POMDP. We denote the agent's observation at time $t$ within an episode as $x_t$, and assume that it maintains an internal state $m_t$, which we can take to be, for example, a recurrent state (such as from an LSTM), or a variable-length array of embeddings of previous observations $x_{0:t}$ (as used in a Transformer). We denote the agent's state update function as $f_\theta$, with learned parameters, $\theta$, s.t.\ $m_{t+1} = f_{\theta}(x_{t+1}, m_t)$. We denote its policy as $\pi_t = h_\theta(m_t)$.

\begin{figure}[t]
\centering
\includegraphics[width=0.87\textwidth]{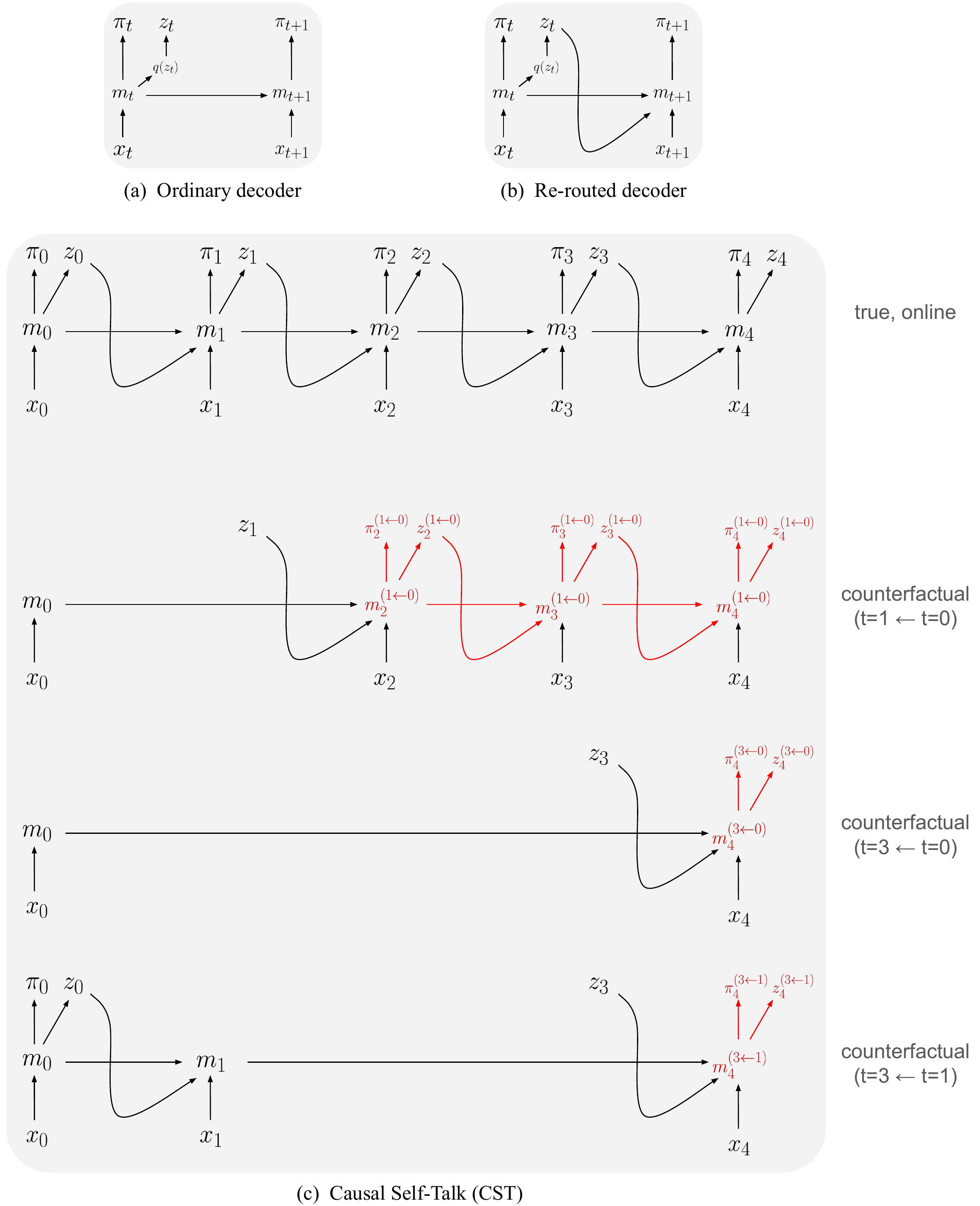}
\caption{
\centering
\textbf{Architectures.}
\textbf{(a)}
Decoders offer a common and simple solution to modeling a base system, albeit one that lacks faithfulness.
\textbf{(b)}
Re-routing the output of a decoder back into the base system is not sufficient to remedy this.
\textbf{(c) The CST architecture} expands on the re-routed decoder by also allowing memory states to be reverted to earlier states before applying a state update. For example, in the second row, the memory state $m_1$ is reverted to $m_0$, so that the (true, online) message $z_1$ must inform the agent about changes to its current state and/or required future policy. Similarly, in the final row, the memory state $m_4$ is reverted to $m_1$, so that the (true, online) message $z_3$ must inform the agent accordingly. Black denotes variables from the true, online rollout of the agent.
Red denotes variables after a counterfactual intervention. The variant of CST used (CST-RL/MR/PD) determines how the counterfactual changes are produced and trained.
}
\label{fig:schematic}
\end{figure}

Our construction starts with a decoder, shown in Figure~\ref{fig:schematic}a. A decoder maps from the state $m_t$ to an output $z_t \sim q(z_t)$, via a learned function $g_{\theta}: m_t \rightarrow q(z_t)$. This mapping is trained using target data $\hat{z} \in \mathcal{D}_z$ that lives in a semantically-meaningful space. As we previously identified in Section~\ref{subsection-non-solutions}, decoders benefit from being grounded, flexible, scalable, and minimally-interfering. However, their weakness is faithfulness. This is because the output $z_t$ merely reflects what information is present in $m_t$, and not whether that information is actually used in the computation of $\pi_t$.

\paragraph{Defining faithfulness.} We now chart a path forward from decoders, with an aim of making them faithful. To do so, we first define two metrics for faithfulness, one weak, one strong:
 
\begin{changemargin}{0.5cm}{0.5cm}
\textbf{Correlational faithfulness:  } if the decoder reports a value of $z_t = z$, to what degree is the behavior (from the policy $\pi$) \textit{consistent} with this value, $z$?
\end{changemargin}

\begin{changemargin}{0.5cm}{0.5cm}
\textbf{Causal faithfulness:\protect\footnote{Our use of this term is distinct from the ``causal faithfulness condition'' in causal modelling \citep{Spirtes2000-xx}.}  } if we were to counterfactually \textit{intervene} on $z_t$, changing it to a value of $z^\prime$, to what degree would this produce behavior consistent with this new value, $z^\prime$, rather than its original value, $z$?
\end{changemargin}

Causal faithfulness is a much stronger property than correlational faithfulness. It's highly desirable because it allows one to validate the agent's commitment to the value of $z_t$ at runtime by making interventions on it. This validation mechanism is especially useful when out of distribution. 

\paragraph{Re-routed decoders.} While it is possible for a decoder to have a degree of correlational faithfulness, it is structurally impossible for it to exhibit causal faithfulness. A simple fix is to route the output of the decoder back into the internal representation (Figure~\ref{fig:schematic}b), e.g.\ by concatenating $z_t$ with $x_{t+1}$. This amounts to incorporating the decoder output in the state update rule, with $m_{t+1} = f_{\theta}([x_{t+1}, z_t], m_t)$.

The difficulty with this simple solution is that there is no guarantee that $f_\theta$ will actually "listen" to $z_t$ when producing $m_{t+1}$. This is because the information in $z_t$ is largely if not completely redundant with $m_t$. This may be further exacerbated by the presence of any stochastic sampling in the generation of $z_t$ from $m_t$. Indeed, as shall be described in Section~\ref{section-results}, our experiments find that re-routed decoders make no measurable progress beyond ordinary decoders on causal faithfulness.

How do we render the function $f_\theta$ sensitive to the value of $z$, when there is redundancy between $z_t$ and $m_t$? It is useful to frame the problem in terms of communication. The decoder pathway can be viewed as a (externally-grounded) communication channel between the internal state and itself, i.e.\ a form of "self-talk" or "inner speech" \citep{Vygotsky2012-se}. The problem therein is that the speaker ($m_t$) and the listener (also $m_t$) are identical, so there is no useful information in the message $z_t$ for the listener.

\paragraph{Causal self-talk.} To overcome this problem, we need to create an information asymmetry between speaker and listener. We introduce \textbf{\textit{Causal Self-Talk} (CST)}: we encourage $f_\theta$ to be sensitive to the message $z_t$ by creating an augmented data distribution where the speaker and listener diverge. Fortunately, we have a ready source of alternate listeners: internal states $m_{t^\prime}$ for $t^\prime < t$.

We thus provide the decoder output with an additional role: to compactly and effectively communicate information to \textit{earlier versions of the same agent} (from the same episode), via a semantically-grounded communication channel. 
Formally, this takes the form of an intervention (Fig~\ref{fig:schematic}c): the internal state, $m_t$, is replaced with a counterfactual one, $m_{t^\prime}$, before being integrated with the original message ($z_t$) and next observation ($x_{t+1}$), in order to produce a counterfactual internal state, $m_{t+1}^{(t \leftarrow t^\prime)} = f_{\theta}([x_{t+1}, z_t], m_{t^\prime})$. This yields a counterfactual policy, $\pi_{t+1}^{(t \leftarrow t^\prime)} = h_{\theta}(m_{t+1}^{(t \leftarrow t^\prime)})$.

This alone, however, is insufficient to fully specify a CST algorithm. We must also stipulate the desired outcome from enacting these interventions. We describe three alternative choices:

\begin{changemargin}{0.5cm}{0.5cm}
\textbf{CST-RL:  } The messages $z_t$ should allow earlier versions of the same agent to maximize return given the current environment state. To train this, interventions must occur online (according to some schedule), and we must allow the agent to act according to the counterfactual policy following an intervention. We then use the maximization of subsequent discounted reward as the desired outcome from the intervention, via reinforcement learning (RL).
\end{changemargin}

\begin{changemargin}{0.5cm}{0.5cm}
\textbf{CST-MR:  } The messages $z_t$ should allow earlier versions of the same agent to reconstruct the next internal state, $m_{t+1}$. Unlike CST-RL, this can be done entirely in replay, without requiring the agent to ever act according to the counterfactual policy on a live environment. We add a weighted memory reconstruction (MR) term to the overall loss: $\mathcal{L}_{MR} = ||m_{t+1}^{(t \leftarrow t^\prime)} - m_{t+1}||^2$.
\end{changemargin}

\begin{changemargin}{0.5cm}{0.5cm}
\textbf{CST-PD:  } The messages $z_t$ should allow earlier versions of the same agent to recover the true current (and future) policy. As with CST-MR, interventions can be simulated entirely in replay. We add a weighted policy distillation (PD) term to the overall loss, using discounting function $\gamma(\cdot)$:
$\mathcal{L}_{PD} \;=\; \sum_{\Delta t > 0}
  \gamma(\Delta t)
  \cdot
  D_{KL}
  \left( \,
  \pi_{t + \Delta t} \, || \,
  \pi_{t + \Delta t}^{(t \leftarrow t^\prime)}
  \, \right)$.
\end{changemargin}

\paragraph{CST and the five desiderata.} 
CST offers a different, less extreme tradeoff between the desiderata. By design, it inherits decoders' resolution to several desiderata.
Like decoders, CST can achieve grounding by supervising the representations $z$ against a signal of known semantics. It is also equally flexible.
Relative to decoders, there is some trade-off with minimal-interference as the decoder output $z_t$ is now an obligate input into the next time step. However, much like decoders, the effects on training can, in principle, be contained through appropriate gradient flow, e.g.\ by limiting updates to parameters of the state update function, $f_\theta$, and/or fine-tuning a pre-existing base system.

CST is highly scalable. CST-RL is straightforward to implement; in its simplest incarnation, fixing $t^\prime=0$ reduces to dropout on the whole internal state, $m_t$. The use of RL, however, occurs at the expense of additional data as it requires online execution of the counterfactual policy. CST-PD and CST-MR, in contrast, are entirely self-supervised: they make use of the existing trajectories generated during the training of the base system. These methods augment the data in a manner similar to the interchange intervention technique of \cite{Geiger2021-cd}, using self-imitation as the desired outcome of the intervention. Depending on the degree of non-stationarity afforded by the particular environment (and its degree of partial observability), the size of the augmented dataset can scale quadratically with the length of each episode (from choosing both $t$ and $t^\prime$); in some settings, it may be possible to exceed this by sampling counterfactual previous states $m_{t^\prime}$ from other episodes.

Finally, unlike decoders, CST is explicitly trained to provide causal faithfulness, by facilitating counterfactual control. Our goal in the following two sections is to evaluate how well it does this.

\section{Experiments}
\label{section-experiments}

\paragraph{Task.}

We study variants of CST in a 3D virtual environment built in Unity \citep{Ward2020-ls, Abramson2020-pa}. This features a fixed-layout indoor space comprising five rooms, each with a different wall color.
The agent receives as input a \nth{1}-person visual observation and a text observation.
The agent can freely navigate using a 4D continuous action space ($a \in [-1, 1]^4$) consisting of in-plane movements and a 2-DOF rotation.

\begin{figure}[t]
\centering
\includegraphics[width=0.9\textwidth]{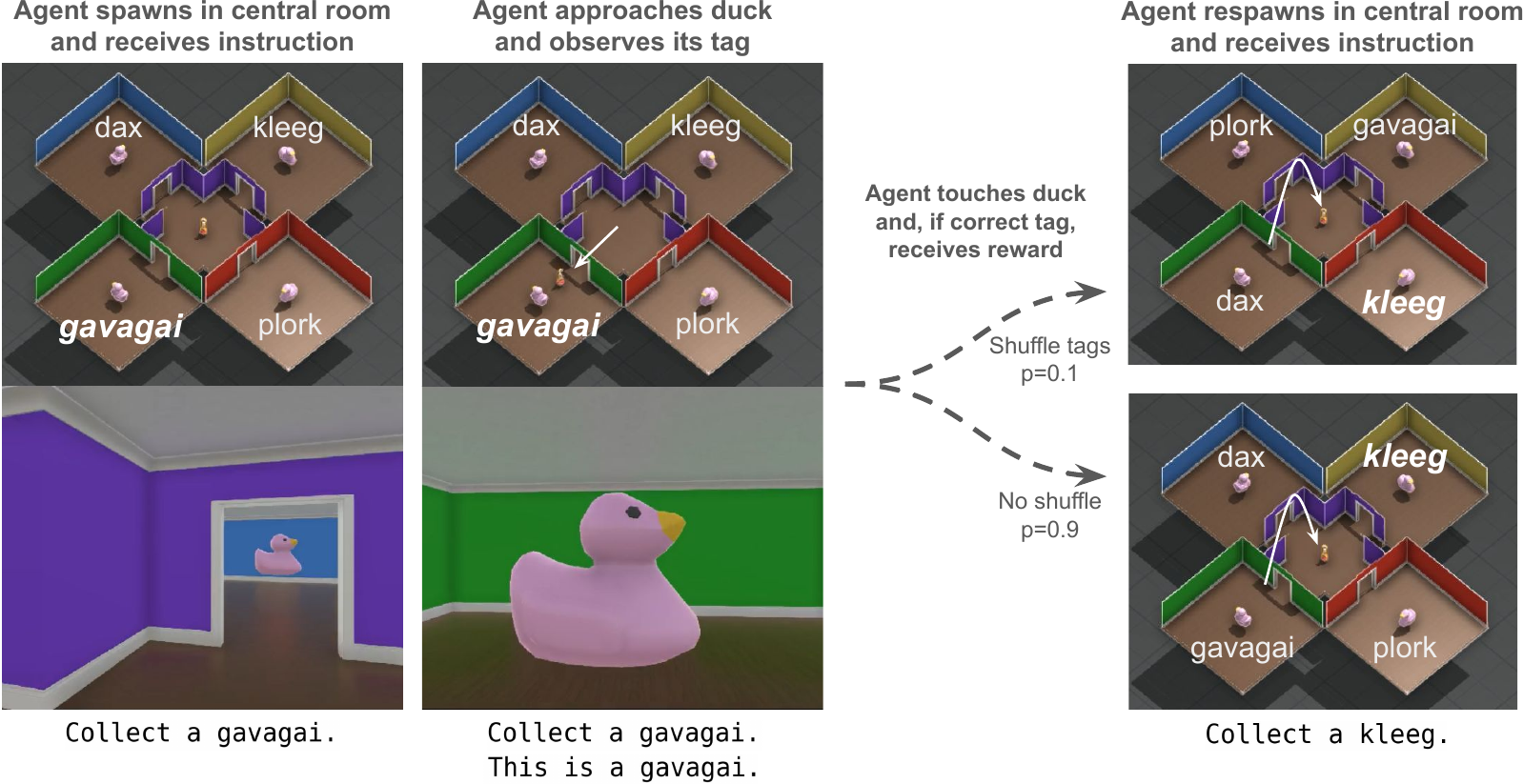}
\caption{
\centering
The \textbf{DaxDucks} task. Top-down views are for exposition only.}
\label{fig:task}
\end{figure}

We developed a simple task in this environment, called \textbf{DaxDucks} (Figure~\ref{fig:task}). We designed DaxDucks to: (1) be easy for an agent to learn a good policy; (2) require an agent to maintain and update beliefs about the latent environment state in order to maximize reward; (3) provide a source of data to ground the self-model; and (4) provide a means to evaluate the faithfulness of the learned self-model.

DaxDucks is an instruction-based, fast-binding search task, where the agent is rewarded for finding a duck that exhibits an instructed tag. Each episode consists of a sequence of trials. At the beginning of each trial, the agent's avatar spawns in the center of the middle room facing a random direction.
Four identical ducks spawn at the center of the four outer rooms. Each duck is randomly assigned one of four tags (``\texttt{dax}'', ``\texttt{gavagai}'', ``\texttt{kleeg}'', or ``\texttt{plork}'').
 The agent is instructed to collect (i.e. collide with) the duck with a specified tag via the text observation channel, e.g. ``\texttt{Collect a gavagai.}'' 
 The agent may observe a duck's tag by entering the duck's room and orienting the center of its view towards the duck. This adds an additional string to the text observation, e.g. ``\texttt{This is a kleeg.}''.
 Colliding with \textit{any} duck terminates the current trial, in which case the agent is respawned in the center of the middle room again and a random new instruction is delivered; a reward is also delivered if the agent collided with the correct duck.
 Finally, when a new trial starts, the ducks usually maintain their tags, but with a small probability
($p_{sh} = 0.1$), the tags are randomly shuffled between the ducks. Thus the agent can only maximize reward by effectively retaining a belief state over which tag belongs to which duck. Each episode lasts for $5000$ steps regardless of the number of trials completed.

\paragraph{Agent architecture.}

For the base system, we used a standard model-free agent (Appendix~\ref{appendix-architecture}): inputs encoded with a ResNet (images) and LSTM (text); a LSTM memory module; MLPs for policy and value heads; and trained with V-trace \citep{Espeholt2018-ir}. All agents achieved a similarly high return.

We demonstrate the flexibility of grounding by considering two forms for the self-model. In the \textbf{one-hot form}, the message $z_t$ comprises four one-hot vectors (one per tag), expressing which room the agent believes each tag to be in. In the \textbf{language form}, the message $z_t$ is a synthetic language string, expressing which room the agent believes a single chosen tag to be in.

For the one-hot form (Section~\ref{subsection:results:one-hot}), we parameterize $q(z_t) = \prod_{\tau=1}^4 q(z_t^\tau)$ as a product of independent categorical distributions $q(z_t^\tau)$ for the four tags $\tau$, with $g_\theta$ (an MLP) outputting the parameters of each $q(z_t^\tau)$ at time $t$, and a sample $z \sim q(z_t)$ concatenated with the encoded input at the next time step. For the language form (Section~\ref{subsection:results:language}), we use an LSTM for $g_\theta$, sampling $z_t$ directly. Examples from a single trajectory are shown in Figure~\ref{fig:one-hot:examples} (one-hot) and Figure~\ref{fig:lang:examples} (language). 
We ground both forms of self-model by supervising $z_t$ against ground-truth values. For the language form, this amounts to constructing a target $\hat{z}_t$ of the form: 
 ``\texttt{The <instructed tag> is in the <color> room.}''.

Given the simplicity of the environment, we implement all three CST algorithms using $t^\prime = 0$ throughout. This means that CST aims to drive the decoder to communicate its beliefs to the earlier version of the agent from the start of the episode. Although this reduces the scale of data augmentation from quadratic to linear, and requires interventions on $z$ to be accompanied by a reset of the memory state to $m_0$, in return it makes it easier to measure the effects of interventions on the agent's behavior.

We used the following schedules for interventions at training time. For CST-RL, interventions $(m_t \leftarrow m_0)$ occurred randomly, with probability $p=0.03$ that an intervention would occur at any given timestep $t$. For CST-MR, we simulated interventions in replay at every timestep. For CST-PD, we simulated interventions in replay: we divided every trajectory into a sequence of blocks of variable duration, with a $p=0.03$ probability that a new block would start at any time $t$; we computed $\mathcal{L}_{PD}$ only up to a horizon of the end of the block, and with a constant discounting function, $\gamma(\Delta t) = 1$.

\section{Results}
\label{section-results}

\paragraph{One-hot self-talk.}
\label{subsection:results:one-hot}

\begin{figure}[t]
\centering
\begin{subfigure}[h]{0.69\textwidth}
    \includegraphics[width=\textwidth]{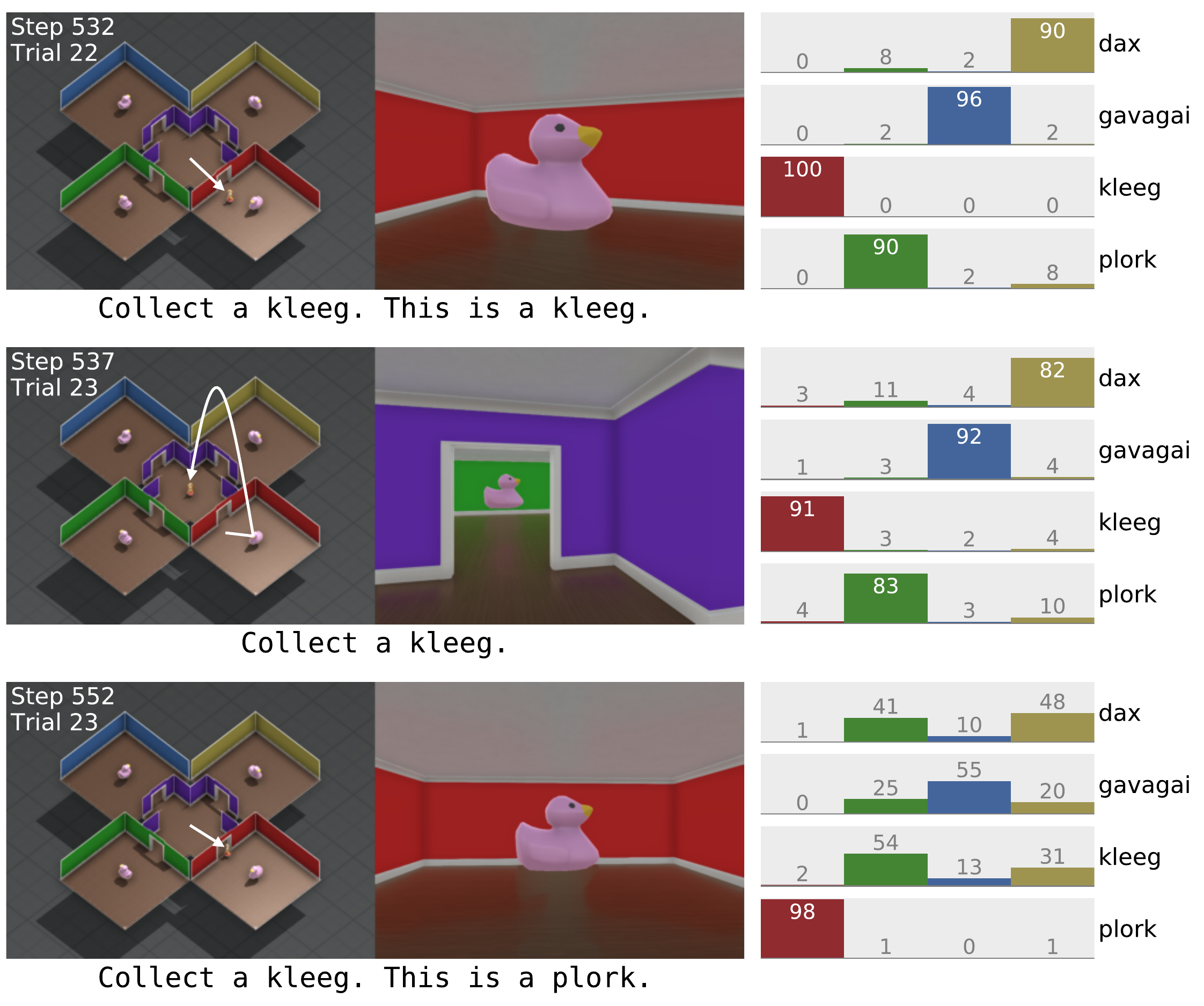}
    \caption{Example one-hot messages from a CST-RL agent.}
    \label{fig:one-hot:examples}
\end{subfigure}
\hspace{0.5cm}
\begin{subfigure}[h]{0.262\textwidth}
    \includegraphics[width=\textwidth]{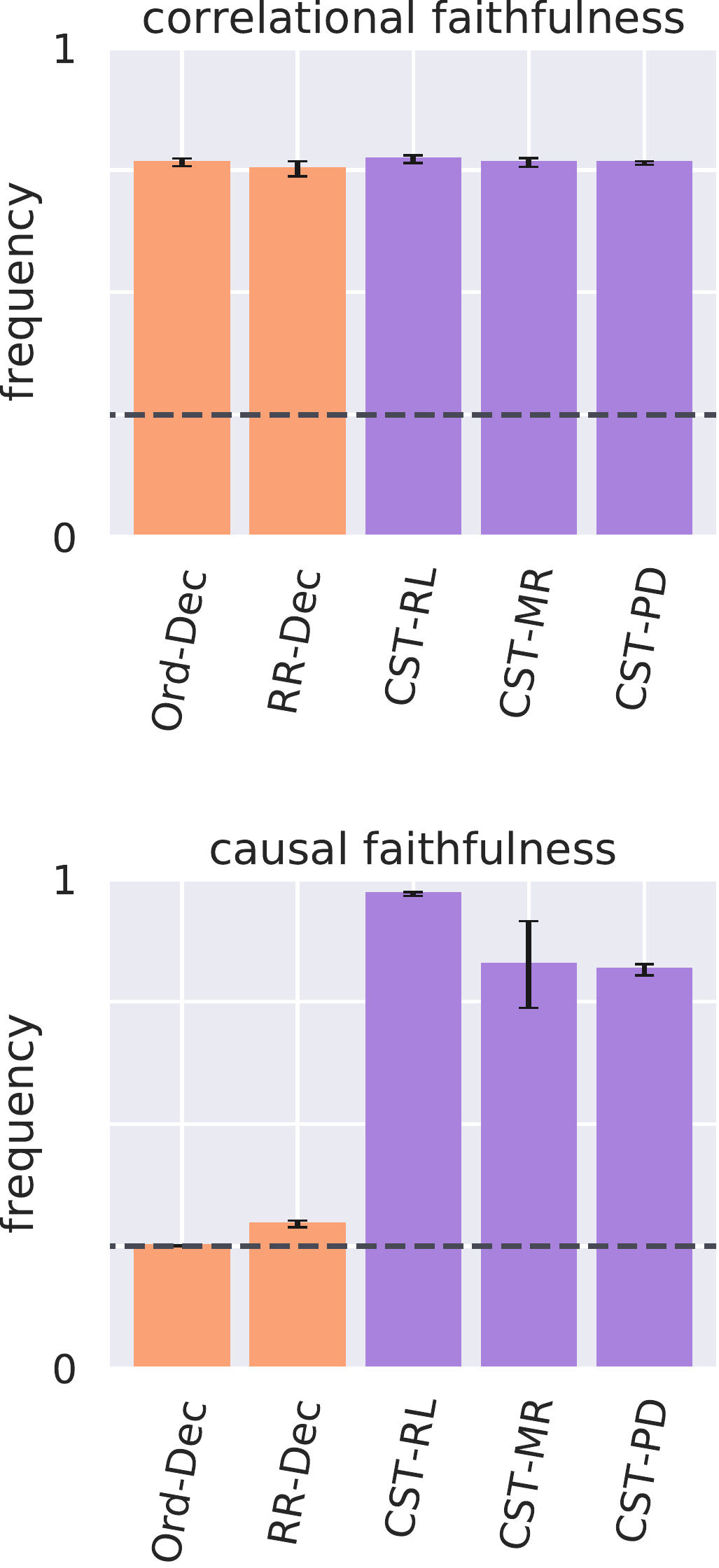}
    \caption{Faithfulness results.}
    \label{fig:one-hot:stats}
\end{subfigure}
\caption{
\centering
\textbf{One-hot-based self-talk.}
\textbf{(a) Example messages during a trajectory.} Top-down view (left), agent view (middle), and $q(z_t)$ (right) during a trajectory of DaxDucks, showing the updating of $q(z_t)$ after the start of a new trial (second row) and after discovering the tags have been shuffled (third row). At each time point, the message $z_t$ is sampled from $q(z_t)$. 
\textbf{(b) CST renders the decoder causally faithful.}  Average values across $5\leq N \leq 10$ training runs per method; error bars show 95\% confidence intervals. Dashed lines indicate faithfulness expected from random messages.
}
\label{fig:one-hot}
\end{figure}

We start by considering the one-hot form for the messages $z$ (Figure~\ref{fig:one-hot:examples}). We compare CST against two baselines: the ordinary decoder (\textbf{Ord-Dec}; Figure~\ref{fig:schematic}a) and the re-routed decoder (\textbf{RR-Dec}; Figure~\ref{fig:schematic}b). We focus on the metrics of faithfulness previously introduced.

\begin{changemargin}{0.5cm}{0.5cm}
\textbf{Correlational faithfulness.  } 
\label{subsection:results:correlational-faithfulness}
When no interventions are taking place, is the agent's physical behavior congruent with the output of its self-model? To measure this, we filter to all times in evaluation episodes when the agent's avatar is in the central room. We then determine the degree to which the belief attested in $q(z_t)$ matches the \textit{next-visited} room, $r$ (whether or not the instructed tag is actually in $r$). For the one-hot message form, this amounts to reading off the value of $q(z_t^\tau = r)$ for the instructed tag $\tau$. 
\end{changemargin}

\begin{changemargin}{0.5cm}{0.5cm}
\textbf{Causal faithfulness.  } 
\label{subsection:results:causal-faithfulness}
When we intervene on $z_t$, is the agent's subsequent physical behavior congruent with this new value? To measure this, we run evaluation episodes where, at the start of each trial, we inject a message $z^\prime$ indicating that the instructed tag is in some random room $r^\prime$. We then measure the probability that the agent actually visits room $r^\prime$ next.
\end{changemargin}

CST maintains comparable levels of correlational faithfulness as the baseline decoders. However, the decoder baselines failed to exhibit a meaningful degree of causal faithfulness. This is indeed impossible for \textbf{Ord-Dec} given the lack of any causal pathway from $z_t$ to $m_{t+1}$. Re-routing $z$ back to the internal state in \textbf{RR-Dec} does result in a small increase above chance, but the effect is relatively tiny.
In contrast, all CST objectives substantially increase the degree of causal faithfulness (Figure~\ref{fig:one-hot:stats}). 

\paragraph{Language-based self-talk.}
\label{subsection:results:language}

\begin{figure}[t]
\centering
\begin{subfigure}[h]{0.67\textwidth}
    \includegraphics[width=\textwidth]{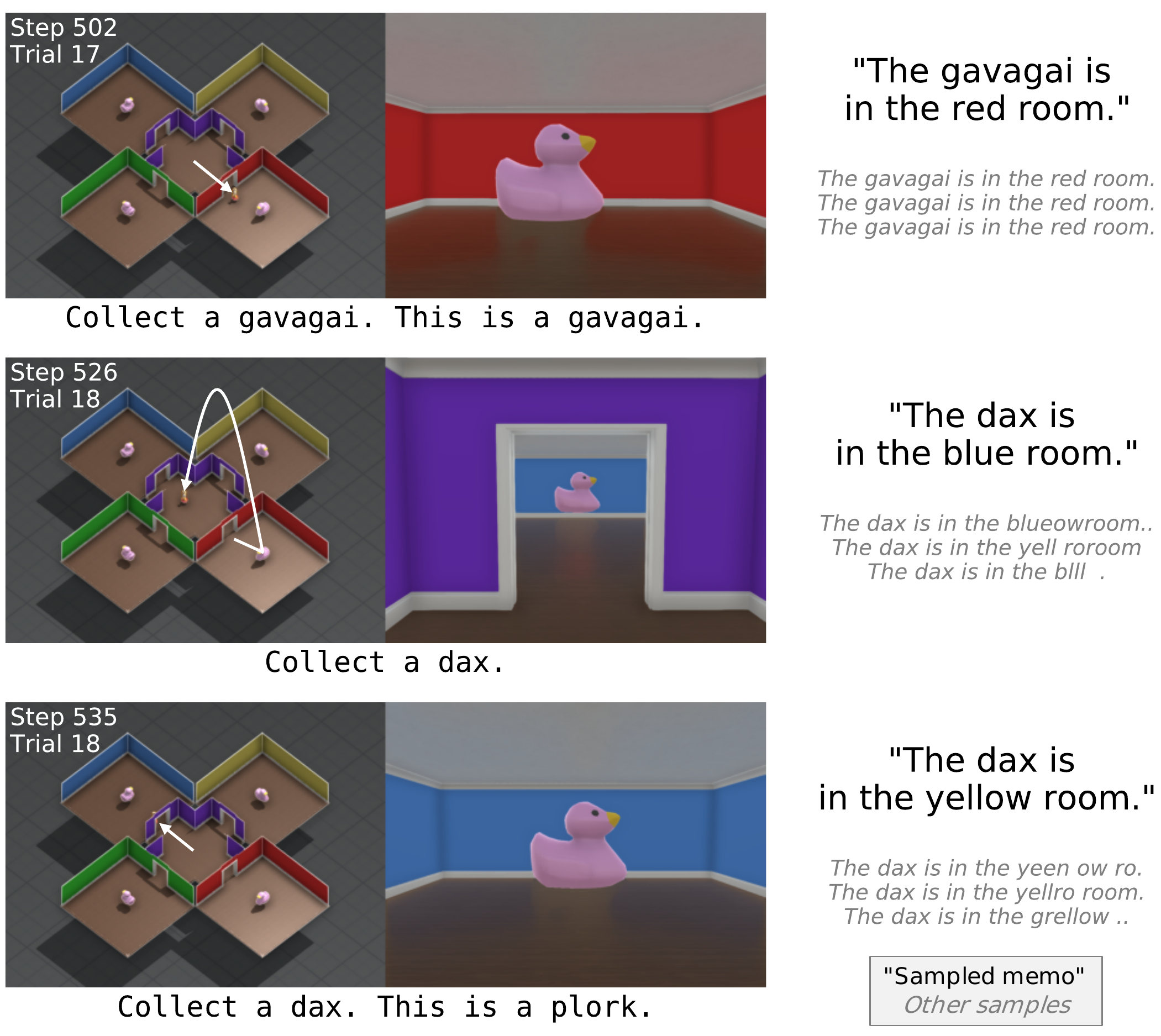}
    \caption{Example language messages from a CST-RL agent.}
    \label{fig:lang:examples}
\end{subfigure}
\hspace{0.5cm}
\begin{subfigure}[h]{0.275\textwidth}
    \includegraphics[width=\textwidth]{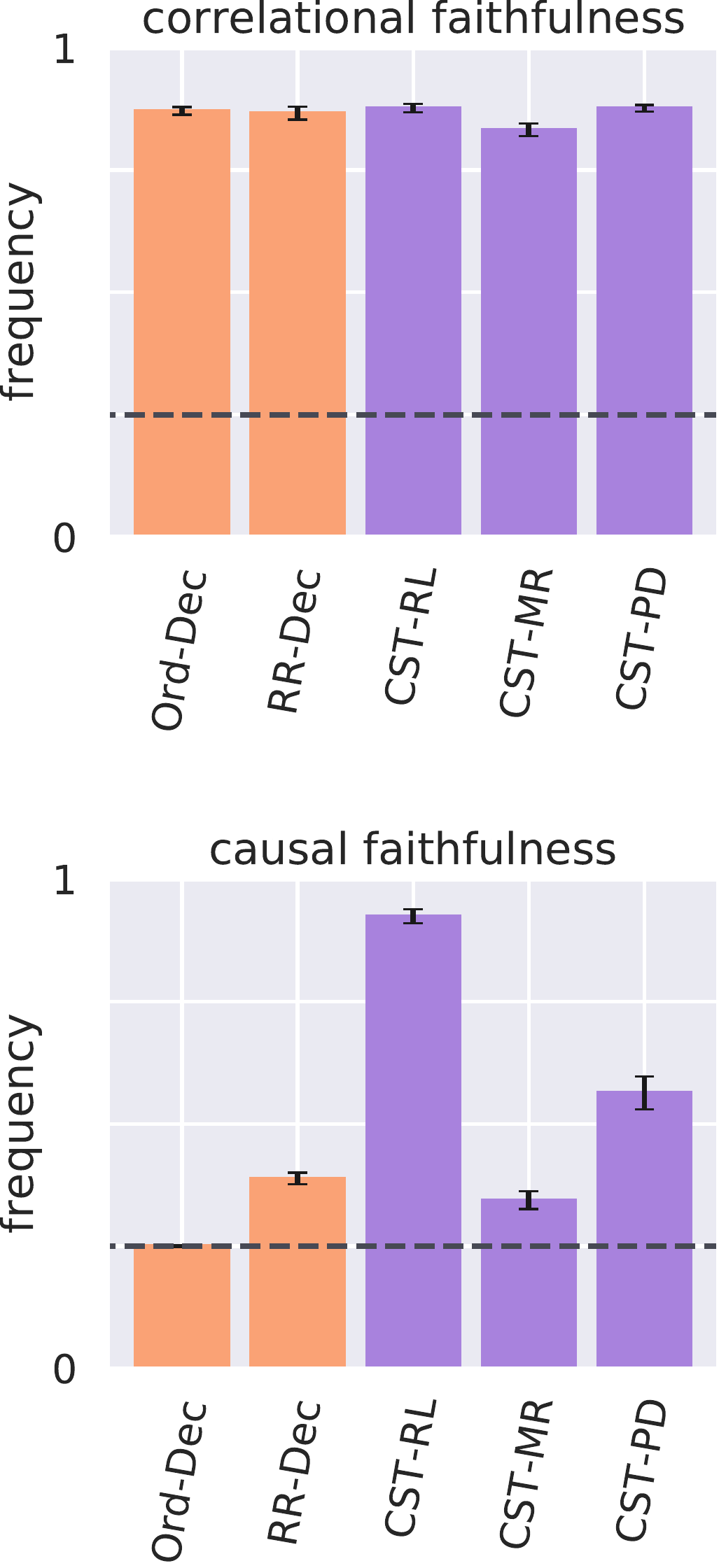}
    \caption{Faithfulness results.}
    \label{fig:lang:stats}
\end{subfigure}
\caption{
\centering
\textbf{Language-based self-talk.} Results as in Figure~\ref{fig:one-hot}.
}
\label{fig:message-examples}
\end{figure}

Next, we trained agents to model their belief state using messages $z$ in a synthetic language form (e.g.\ Figure~\ref{fig:lang:examples}).
For correlational faithfulness, we obtain analogous values of $q(z_t^\tau = r)$ by computing the likelihoods of generating the four messages ``\texttt{The <instructed tag> is in the <color> room.}'', and normalizing these to sum to 1.

As with the one-hot form, we found that the greatest gain of CST was its ability to imbue the self-model with causal faithfulness. The effect was most profound with CST-RL, and moderate with CST-PD. However, we found it difficult to get any traction with CST-MR; this was either disruptive to learning the base policy, or ineffective at causal faithfulness.

We note that the faithfulness results with the language form were overall weaker than for the one-hot form. This may be because the one-hot messages encode a room belief about all four tags.

\begin{figure}[t]
\centering
\includegraphics[width=0.68\textwidth]{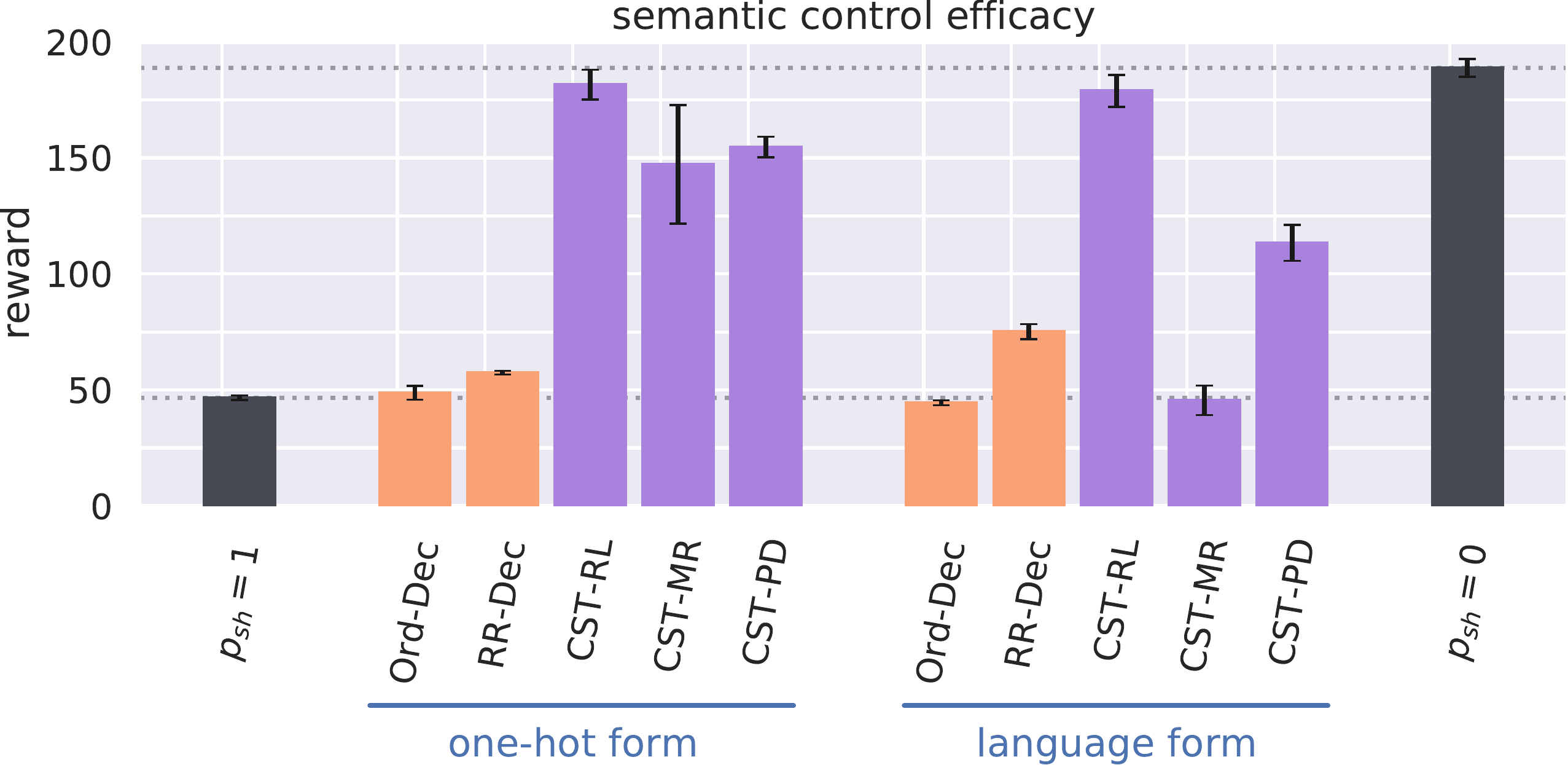}
\caption{\textbf{Efficacy of semantic control via interventions on the self-model.}
\label{fig:semantic-control-efficacy}}
\end{figure}

\paragraph{Semantic control.}
\label{subsection:results:semantic-control}

As a final test, we consider whether CST endows the self-model with sufficient causal power to act as an effective semantic control interface for the agent.
We thus measure how effectively the agent can use injected information to obtain reward. We start with two baselines: (1) the return on evaluation episodes where $p_{sh}=1$, i.e.\ where tags are shuffled on every trial; and (2) the return on evaluation episodes where $p_{sh}=0$, i.e.\ where tags are never shuffled. Respectively, these represent the bounds of rewards the agent naturally obtains when it can acquire (1) minimal or (2) maximal information about the location of the instructed tag. We compare these to the reward obtained on evaluation episodes where information about the tag location is provided through intervening on $z$. Specifically, we set $p_{sh}=1$, and inject a message $z^\prime$ at the start of each trial indicating the location of all the tags (for the one-hot form), or the location of the instructed tag (for the language form). 
The results (Figure~\ref{fig:semantic-control-efficacy}) show that CST not only furnishes an agent with a semantically-grounded model that describes its behavior, but also one which can be used as a mechanism to guide it.
\section{Discussion}
\label{section-discussion}
We articulated five key desiderata of an ideal explanation system for an AI system: grounding, flexibility, minimal-interference, scalability, and faithfulness. Unlike existing approaches, we seek to satisfy all of these by training the base system to serve a causal model of itself. We developed an instance of this solution for Deep RL agents: Causal Self-Talk. In CST, an agent learns to serve a grounded and faithful model of its internal state by learning to communicate with itself across time. We demonstrated its effectiveness under various forms in a simple embodied task in a simulated 3D environment. We also showed that CST enables a new mechanism for effective semantic control.

We presented three variants of the CST algorithm. CST-RL yielded the greatest degree of causal faithfulness and efficacy of semantic control. Here, online interventions at training time force the agent to learn how to use the content of its self-model to maximize its reward. CST-RL's success, however, may reflect features of the particular task setting we study: here, there is a close alignment between writing a faithful $z$ and maximizing reward. This may not always be the case.

In contrast, CST-PD offers an attractive alternative for two reasons: (1) it trains entirely in replay, and thus potentially interferes less with the base system; and (2) as it primarily seeks to enable \textit{imitation} of the current behavior, the faithfulness of the self-model is targeted by design. One challenge here is learning to capture the important features of behavior; more impressive gains may be possible using richer imitation learning techniques \citep[e.g.][]{Ho2016-uz} rather than behavioral cloning. We had the least success with CST-MR: there also may be theoretical obstacles to deploying CST-MR in some settings, e.g.\ when the dimensionality of the internal state is very large, or is of variable length.

\paragraph{Limitations.}
In our experiments, we use a limited form of CST, where we exclusively revert memory states $m_t$ to the state at time $t^\prime=0$, rather than the more general case of $t^\prime < t$.
We do this to simplify the presentation and evaluation of the technique, but it imparts two limitations: (1) it requires $m_t$ to be reverted to $m_0$ in order to effectively intervene on the message $z$; and (2) it requires $z$ to contain all behaviorally-relevant information. 
While sufficient for the simple task at hand, this becomes unrealistic in more complex environments. 
When $z$ contains only a partial description of $m_t$, it will be necessary for the agent to be able to re-integrate injected $z$ values with its current memory state. The more general case of $t^\prime < t$ offers a viable path to achieving this.

Utility requires that the self-model be grounded in a semantically-meaningful form. We have shown how a ready source of data can be used to train the self-model. How this method stands up in more limited data regimes, or in OOD scenarios where the self-model hasn't been supervised,
remains an open question. It may be possible in some domains to leverage semi-supervised methods, and/or to impose structure on the self-model as a helpful constraint.

\paragraph{Future opportunities.}
We focused here on expressing the agent's beliefs about the current state of the environment in the self-model. A complete answer to the question, "Why did you do that?" might also include an expression of an agent's goals, plans, and model of the world \citep{Miller2019-la}. The schema we have developed here should be sufficiently flexible to extend in this direction, given the right training data. It would also be of great value to determine how to apply these techniques beyond RL agents.

Our primary focus in this paper has been on building self-models to yield explanations.
However, this is only one of several utilities that a faithful self-model might deliver.
For example, we demonstrate that causal self-models provide a semantic control mechanism.
This could be useful for guiding agent learning through human interaction \citep{Abramson2020-pa}, metacognitive reporting and control \citep{Hamrick2017-am, Banino2021-yu, Fleur2021-fr}, or as an interface for safety-critical applications \citep{Matuszek2013-in, Khayrallah2015-hg, Roh2020-zd, Kim2020-wp}.
By learning to express its internal state through a different representational schema, this may also open up avenues for new exploration or hierarchical reinforcement learning techniques.

Finally, training agents to constantly talk to themselves might prove to be a data-efficient way of enabling agents to talk to others (and us!). 
Ultimately, this may pave a way for us to invert of the flow of knowledge, allowing agents to not only answer, "Why did you do that?", but also "How?".

\begin{ack}
The authors would like to thank Edgar Du\'{e}\~{n}ez-Guzm\'{a}n for early environment design, Hamza Merzic for agent infrastructure assistance, Allison Tam for help in agent development, Chen Yan for useful discussion, Lucy Campbell-Gillingham for organizational support, and Johannes Welbl, Kory Mathewson, Pedro Ortega, Agnieszka Grabska-Barwi\'{n}ska, Chen Yan, Matt Botvinick, and Lisa Anne Hendricks for comments on the work and/or the manuscript.
\end{ack}

\bibliography{library}

\begin{thebibliography}{74}
\providecommand{\natexlab}[1]{#1}
\providecommand{\url}[1]{\texttt{#1}}
\expandafter\ifx\csname urlstyle\endcsname\relax
  \providecommand{\doi}[1]{doi: #1}\else
  \providecommand{\doi}{doi: \begingroup \urlstyle{rm}\Url}\fi

\bibitem[Abramson et~al.(2020)Abramson, Ahuja, Barr, Brussee, Carnevale,
  Cassin, Chhaparia, Clark, Damoc, Dudzik, and {Others}]{Abramson2020-pa}
Josh Abramson, Arun Ahuja, Iain Barr, Arthur Brussee, Federico Carnevale, Mary
  Cassin, Rachita Chhaparia, Stephen Clark, Bogdan Damoc, Andrew Dudzik, and
  {Others}.
\newblock Imitating interactive intelligence.
\newblock \emph{arXiv preprint arXiv:2012.05672}, 2020.

\bibitem[Adebayo et~al.(2018)Adebayo, Gilmer, Muelly, Goodfellow, Hardt, and
  Kim]{Adebayo2018-mg}
Julius Adebayo, Justin Gilmer, Michael Muelly, Ian Goodfellow, Moritz Hardt,
  and Been Kim.
\newblock Sanity checks for saliency maps.
\newblock \emph{Adv. Neural Inf. Process. Syst.}, 31, 2018.

\bibitem[Akata et~al.(2018)Akata, Hendricks, Alaniz, and Darrell]{Akata2018-al}
Zeynep Akata, Lisa~Anne Hendricks, Stephan Alaniz, and Trevor Darrell.
\newblock Generating post-hoc rationales of deep visual classification
  decisions.
\newblock In \emph{The Springer Series on Challenges in Machine Learning},
  pages 135--154. Springer International Publishing, Cham, 2018.

\bibitem[Alvarez-Melis and Jaakkola(2018)]{Alvarez-Melis2018-xo}
David Alvarez-Melis and Tommi~S Jaakkola.
\newblock Towards robust interpretability with {Self-Explaining} neural
  networks.
\newblock \emph{arXiv preprint arXiv:1806.07538}, June 2018.

\bibitem[Andreas et~al.(2016)Andreas, Klein, and Levine]{Andreas2016-ao}
Jacob Andreas, Dan Klein, and Sergey Levine.
\newblock Modular multitask reinforcement learning with policy sketches.
\newblock \emph{arXiv preprint arXiv:1611.01796}, November 2016.

\bibitem[Anirudh et~al.(2017)Anirudh, Bremer, Sridhar, and
  Thiagarajan]{Anirudh2017-xx}
Rushil Anirudh, P~Bremer, Rahul Sridhar, and JJ~Thiagarajan.
\newblock Influential sample selection: A graph signal processing approach.
\newblock Technical report, Lawrence Livermore National Lab.(LLNL), Livermore,
  CA (United States), 2017.

\bibitem[Atrey et~al.(2019)Atrey, Clary, and Jensen]{Atrey2019-ma}
Akanksha Atrey, Kaleigh Clary, and David Jensen.
\newblock Exploratory not explanatory: Counterfactual analysis of saliency maps
  for deep reinforcement learning.
\newblock \emph{arXiv preprint arXiv:1912.05743}, 2019.

\bibitem[Bach et~al.(2015)Bach, Binder, Montavon, Klauschen, M{\"u}ller, and
  Samek]{Bach2015-lz}
Sebastian Bach, Alexander Binder, Gr{\'e}goire Montavon, Frederick Klauschen,
  Klaus-Robert M{\"u}ller, and Wojciech Samek.
\newblock On {Pixel-Wise} explanations for {Non-Linear} classifier decisions by
  {Layer-Wise} relevance propagation.
\newblock \emph{PLOS ONE}, 10\penalty0 (7):\penalty0 e0130140, 2015.

\bibitem[Banino et~al.(2021)Banino, Balaguer, and Blundell]{Banino2021-yu}
Andrea Banino, Jan Balaguer, and Charles Blundell.
\newblock Pondernet: Learning to ponder.
\newblock \emph{arXiv preprint arXiv:2107.05407}, 2021.

\bibitem[Battaglia et~al.(2018)Battaglia, Hamrick, Bapst, Sanchez-Gonzalez,
  Zambaldi, Malinowski, Tacchetti, Raposo, Santoro, Faulkner, Gulcehre, Song,
  Ballard, Gilmer, Dahl, Vaswani, Allen, Nash, Langston, Dyer, Heess, Wierstra,
  Kohli, Botvinick, Vinyals, Li, and Pascanu]{Battaglia2018-xw}
Peter~W Battaglia, Jessica~B Hamrick, Victor Bapst, Alvaro Sanchez-Gonzalez,
  Vinicius Zambaldi, Mateusz Malinowski, Andrea Tacchetti, David Raposo, Adam
  Santoro, Ryan Faulkner, Caglar Gulcehre, Francis Song, Andrew Ballard, Justin
  Gilmer, George Dahl, Ashish Vaswani, Kelsey Allen, Charles Nash, Victoria
  Langston, Chris Dyer, Nicolas Heess, Daan Wierstra, Pushmeet Kohli, Matt
  Botvinick, Oriol Vinyals, Yujia Li, and Razvan Pascanu.
\newblock Relational inductive biases, deep learning, and graph networks.
\newblock \emph{arXiv preprint arXiv:1806.01261}, June 2018.

\bibitem[Bau et~al.(2017)Bau, Zhou, Khosla, Oliva, and Torralba]{Bau2017-lx}
David Bau, Bolei Zhou, Aditya Khosla, Aude Oliva, and Antonio Torralba.
\newblock Network dissection: Quantifying interpretability of deep visual
  representations.
\newblock \emph{arXiv preprint arXiv:1704.05796}, April 2017.

\bibitem[Belinkov and Glass(2019)]{Belinkov2019-in}
Yonatan Belinkov and James Glass.
\newblock Analysis methods in neural language processing: A survey.
\newblock \emph{Transactions of the Association for Computational Linguistics},
  7:\penalty0 49--72, 2019.

\bibitem[Biecek and Burzykowski(2021)]{Biecek2021-va}
Przemyslaw Biecek and Tomasz Burzykowski.
\newblock Local interpretable model-agnostic explanations ({LIME}).
\newblock \emph{Explanatory Model Analysis}, pages 107--123, 2021.

\bibitem[Bien and Tibshirani(2011)]{Bien2011-xx}
Jacob Bien and Robert Tibshirani.
\newblock Prototype selection for interpretable classification.
\newblock \emph{The Annals of Applied Statistics}, 5\penalty0 (4):\penalty0
  2403--2424, 2011.

\bibitem[Bonnen et~al.(2021)Bonnen, Yamins, and Wagner]{Bonnen2021-zy}
Tyler Bonnen, Daniel L~K Yamins, and Anthony~D Wagner.
\newblock When the ventral visual stream is not enough: A deep learning account
  of medial temporal lobe involvement in perception.
\newblock \emph{Neuron}, 109\penalty0 (17):\penalty0 2755--2766.e6, September
  2021.

\bibitem[Burgess et~al.(2019)Burgess, Matthey, Watters, Kabra, Higgins,
  Botvinick, and Lerchner]{Burgess2019-aq}
Christopher~P Burgess, Loic Matthey, Nicholas Watters, Rishabh Kabra, Irina
  Higgins, Matt Botvinick, and Alexander Lerchner.
\newblock {MONet}: Unsupervised scene decomposition and representation.
\newblock \emph{arXiv preprint arXiv:1901.11390}, January 2019.

\bibitem[Cammarata et~al.(2020)Cammarata, Goh, Carter, Schubert, Petrov, and
  Olah]{Cammarata2020-my}
Nick Cammarata, Gabriel Goh, Shan Carter, Ludwig Schubert, Michael Petrov, and
  Chris Olah.
\newblock Curve detectors.
\newblock \emph{Distill}, 5\penalty0 (6), June 2020.

\bibitem[Cho et~al.(2020)Cho, Lee, and Choi]{Cho2020-sd}
Hyeoncheol Cho, Eok~Kyun Lee, and Insung~S Choi.
\newblock Layer-wise relevance propagation of {InteractionNet} explains
  protein--ligand interactions at the atom level.
\newblock \emph{Scientific Reports}, 10\penalty0 (1), 2020.

\bibitem[D{\'e}letang et~al.(2021)D{\'e}letang, Grau-Moya, Martic, Genewein,
  McGrath, Mikulik, Kunesch, Legg, and Ortega]{Deletang2021-xy}
Gr{\'e}goire D{\'e}letang, Jordi Grau-Moya, Miljan Martic, Tim Genewein, Tom
  McGrath, Vladimir Mikulik, Markus Kunesch, Shane Legg, and Pedro~A Ortega.
\newblock Causal analysis of agent behavior for ai safety.
\newblock \emph{arXiv preprint arXiv:2103.03938}, 2021.

\bibitem[Elazar et~al.(2021)Elazar, Ravfogel, Jacovi, and
  Goldberg]{Elazar2021-zx}
Yanai Elazar, Shauli Ravfogel, Alon Jacovi, and Yoav Goldberg.
\newblock Amnesic probing: Behavioral explanation with amnesic counterfactuals.
\newblock \emph{Transactions of the Association for Computational Linguistics},
  9:\penalty0 160--175, 2021.

\bibitem[Elhage et~al.(2021)Elhage, Nanda, Olsson, Henighan, Joseph, Mann,
  Askell, Bai, Chen, Conerly, DasSarma, Drain, Ganguli, Hatfield-Dodds,
  Hernandez, Jones, Kernion, Lovitt, Ndousse, Amodei, Brown, Clark, Kaplan,
  McCandlish, and Olah]{Elhage2021-gn}
Nelson Elhage, Neel Nanda, Catherine Olsson, Tom Henighan, Nicholas Joseph, Ben
  Mann, Amanda Askell, Yuntao Bai, Anna Chen, Tom Conerly, Nova DasSarma, Dawn
  Drain, Deep Ganguli, Zac Hatfield-Dodds, Danny Hernandez, Andy Jones, Jackson
  Kernion, Liane Lovitt, Kamal Ndousse, Dario Amodei, Tom Brown, Jack Clark,
  Jared Kaplan, Sam McCandlish, and Chris Olah.
\newblock A mathematical framework for transformer circuits.
\newblock \emph{Transformer Circuits Thread}, 2021.

\bibitem[Espeholt et~al.(2018)Espeholt, Soyer, Munos, Simonyan, Mnih, Ward,
  Doron, Firoiu, Harley, Dunning, and {Others}]{Espeholt2018-ir}
Lasse Espeholt, Hubert Soyer, Remi Munos, Karen Simonyan, Vlad Mnih, Tom Ward,
  Yotam Doron, Vlad Firoiu, Tim Harley, Iain Dunning, and {Others}.
\newblock Impala: Scalable distributed deep-rl with importance weighted
  actor-learner architectures.
\newblock In \emph{International Conference on Machine Learning}, pages
  1407--1416, 2018.

\bibitem[Fauw et~al.(2018)Fauw, De~Fauw, Ledsam, Romera-Paredes, Nikolov,
  Tomasev, Blackwell, Askham, Glorot, O'Donoghue, Visentin, van~den Driessche,
  Lakshminarayanan, Meyer, Mackinder, Bouton, Ayoub, Chopra, King,
  Karthikesalingam, Hughes, Raine, Hughes, Sim, Egan, Tufail, Montgomery,
  Hassabis, Rees, Back, Khaw, Suleyman, Cornebise, Keane, and
  Ronneberger]{Fauw2018-rm}
Jeffrey~De Fauw, Jeffrey De~Fauw, Joseph~R Ledsam, Bernardino Romera-Paredes,
  Stanislav Nikolov, Nenad Tomasev, Sam Blackwell, Harry Askham, Xavier Glorot,
  Brendan O'Donoghue, Daniel Visentin, George van~den Driessche, Balaji
  Lakshminarayanan, Clemens Meyer, Faith Mackinder, Simon Bouton, Kareem Ayoub,
  Reena Chopra, Dominic King, Alan Karthikesalingam, C{\'\i}an~O Hughes,
  Rosalind Raine, Julian Hughes, Dawn~A Sim, Catherine Egan, Adnan Tufail, Hugh
  Montgomery, Demis Hassabis, Geraint Rees, Trevor Back, Peng~T Khaw, Mustafa
  Suleyman, Julien Cornebise, Pearse~A Keane, and Olaf Ronneberger.
\newblock Clinically applicable deep learning for diagnosis and referral in
  retinal disease.
\newblock \emph{Nature Medicine}, 24\penalty0 (9):\penalty0 1342--1350, 2018.

\bibitem[Fleur et~al.(2021)Fleur, Bredeweg, and van~den Bos]{Fleur2021-fr}
Damien~S Fleur, Bert Bredeweg, and Wouter van~den Bos.
\newblock Metacognition: ideas and insights from neuro- and educational
  sciences.
\newblock \emph{NPJ Sci Learn}, 6\penalty0 (1):\penalty0 13, June 2021.

\bibitem[Geiger et~al.(2021)Geiger, Wu, Lu, Rozner, Kreiss, Icard, Goodman, and
  Potts]{Geiger2021-cd}
Atticus Geiger, Zhengxuan Wu, Hanson Lu, Josh Rozner, Elisa Kreiss, Thomas
  Icard, Noah~D Goodman, and Christopher Potts.
\newblock Inducing causal structure for interpretable neural networks.
\newblock \emph{arXiv preprint arXiv:2112.00826}, 2021.

\bibitem[Goh et~al.(2021)Goh, †, †, Carter, Petrov, Schubert, Radford, and
  Olah]{Goh2021-fw}
Gabriel Goh, Nick~Cammarata †, Chelsea~Voss †, Shan Carter, Michael Petrov,
  Ludwig Schubert, Alec Radford, and Chris Olah.
\newblock Multimodal neurons in artificial neural networks.
\newblock \emph{Distill}, 2021.

\bibitem[Goyal et~al.(2019)Goyal, Wu, Ernst, Batra, Parikh, and
  Lee]{Goyal2019-ev}
Yash Goyal, Ziyan Wu, Jan Ernst, Dhruv Batra, Devi Parikh, and Stefan Lee.
\newblock Counterfactual visual explanations.
\newblock In \emph{International Conference on Machine Learning}, pages
  2376--2384, 2019.

\bibitem[Guidotti et~al.(2018)Guidotti, Monreale, Ruggieri, Turini, Giannotti,
  and Pedreschi]{Guidotti2018-cf}
Riccardo Guidotti, Anna Monreale, Salvatore Ruggieri, Franco Turini, Fosca
  Giannotti, and Dino Pedreschi.
\newblock A survey of methods for explaining black box models.
\newblock \emph{ACM computing surveys (CSUR)}, 51\penalty0 (5):\penalty0 1--42,
  2018.

\bibitem[Hamrick et~al.(2017)Hamrick, Ballard, Pascanu, Vinyals, Heess, and
  Battaglia]{Hamrick2017-am}
Jessica~B Hamrick, Andrew~J Ballard, Razvan Pascanu, Oriol Vinyals, Nicolas
  Heess, and Peter~W Battaglia.
\newblock Metacontrol for adaptive imagination-based optimization.
\newblock \emph{arXiv preprint arXiv:1705.02670}, 2017.

\bibitem[Hayes and Shah(2017)]{Hayes2017-gk}
Bradley Hayes and Julie~A Shah.
\newblock Improving robot controller transparency through autonomous policy
  explanation.
\newblock In \emph{Proceedings of the 2017 {ACM/IEEE} International Conference
  on {Human-Robot} Interaction}, New York, NY, USA, March 2017. ACM.

\bibitem[Hendricks et~al.(2016)Hendricks, Akata, Rohrbach, Donahue, Schiele,
  and Darrell]{Hendricks2016-pa}
Lisa~Anne Hendricks, Zeynep Akata, Marcus Rohrbach, Jeff Donahue, Bernt
  Schiele, and Trevor Darrell.
\newblock Generating visual explanations.
\newblock In \emph{European conference on computer vision}, pages 3--19, 2016.

\bibitem[Hendricks et~al.(2018)Hendricks, Hu, Darrell, and
  Akata]{Hendricks2018-af}
Lisa~Anne Hendricks, Ronghang Hu, Trevor Darrell, and Zeynep Akata.
\newblock Generating counterfactual explanations with natural language.
\newblock \emph{arXiv preprint arXiv:1806.09809}, 2018.

\bibitem[Hendricks et~al.(2021)Hendricks, Rohrbach, Schiele, Darrell, and
  Akata]{Hendricks2021-af}
Lisa~Anne Hendricks, Anna Rohrbach, Bernt Schiele, Trevor Darrell, and Zeynep
  Akata.
\newblock Generating visual explanations with natural language.
\newblock \emph{Applied AI Letters}, 2\penalty0 (4), December 2021.

\bibitem[Ho and Ermon(2016)]{Ho2016-uz}
Jonathan Ho and Stefano Ermon.
\newblock Generative adversarial imitation learning.
\newblock \emph{Adv. Neural Inf. Process. Syst.}, 29, 2016.

\bibitem[Jacovi and Goldberg(2020)]{Jacovi2020-hd}
Alon Jacovi and Yoav Goldberg.
\newblock Towards faithfully interpretable {NLP} systems: How should we define
  and evaluate faithfulness?
\newblock In \emph{Proceedings of the 58th Annual Meeting of the Association
  for Computational Linguistics}, pages 4198--4205, Online, July 2020.
  Association for Computational Linguistics.

\bibitem[Jacovi et~al.(2021)Jacovi, Marasovi{\'c}, Miller, and
  Goldberg]{Jacovi2021-nl}
Alon Jacovi, Ana Marasovi{\'c}, Tim Miller, and Yoav Goldberg.
\newblock Formalizing trust in artificial intelligence: Prerequisites, causes
  and goals of human trust in {AI}.
\newblock In \emph{Proceedings of the 2021 {ACM} conference on fairness,
  accountability, and transparency}, pages 624--635, 2021.

\bibitem[Jaderberg et~al.(2017)Jaderberg, Mnih, Czarnecki, Schaul, Leibo,
  Silver, and Kavukcuoglu]{Jaderberg2017-pw}
Max Jaderberg, Volodymyr Mnih, Wojciech~Marian Czarnecki, Tom Schaul, Joel~Z
  Leibo, David Silver, and Koray Kavukcuoglu.
\newblock Reinforcement learning with unsupervised auxiliary tasks.
\newblock In \emph{5th International Conference on Learning Representations,
  {ICLR} 2017, Toulon, France, April 24-26, 2017, Conference Track
  Proceedings}. OpenReview.net, 2017.

\bibitem[Jain and Wallace(2019)]{Jain2019-wt}
Sarthak Jain and Byron~C Wallace.
\newblock Attention is not explanation.
\newblock \emph{arXiv preprint arXiv:1902.10186}, February 2019.

\bibitem[Juozapaitis et~al.(2019)Juozapaitis, Koul, Fern, Erwig, and
  Doshi-Velez]{Juozapaitis2019-ag}
Zoe Juozapaitis, Anurag Koul, Alan Fern, Martin Erwig, and Finale Doshi-Velez.
\newblock Explainable reinforcement learning via reward decomposition.
\newblock In \emph{{IJCAI/ECAI} Workshop on explainable artificial
  intelligence}, 2019.

\bibitem[Kartal et~al.(2019)Kartal, Hernandez-Leal, and Taylor]{Kartal2019-mj}
Bilal Kartal, Pablo Hernandez-Leal, and Matthew~E Taylor.
\newblock Terminal prediction as an auxiliary task for deep reinforcement
  learning.
\newblock In \emph{Proceedings of the Fifteenth {AAAI} Conference on Artificial
  Intelligence and Interactive Digital Entertainment}, AIIDE'19, Atlanta,
  Georgia, 2019. AAAI Press.

\bibitem[Khayrallah et~al.(2015)Khayrallah, Trott, and
  Feldman]{Khayrallah2015-hg}
Huda Khayrallah, Sean Trott, and Jerome Feldman.
\newblock Natural language for human robot interaction.
\newblock In \emph{International Conference on {Human-Robot} Interaction
  ({HRI})}, 2015.

\bibitem[Kim et~al.(2014)Kim, Rudin, and Shah]{Kim2014-xx}
Been Kim, Cynthia Rudin, and Julie~A Shah.
\newblock The bayesian case model: A generative approach for case-based
  reasoning and prototype classification.
\newblock \emph{Advances in neural information processing systems}, 27, 2014.

\bibitem[Kim et~al.(2016)Kim, Khanna, and Koyejo]{Kim2016-xx}
Been Kim, Rajiv Khanna, and Oluwasanmi~O Koyejo.
\newblock Examples are not enough, learn to criticize! criticism for
  interpretability.
\newblock \emph{Advances in neural information processing systems}, 29, 2016.

\bibitem[Kim et~al.(2017)Kim, Wattenberg, Gilmer, Cai, Wexler, Viegas, and
  Sayres]{Kim2017-dr}
Been Kim, Martin Wattenberg, Justin Gilmer, Carrie Cai, James Wexler, Fernanda
  Viegas, and Rory Sayres.
\newblock Interpretability beyond feature attribution: Quantitative testing
  with concept activation vectors ({TCAV}).
\newblock \emph{arXiv preprint arXiv:1711.11279}, November 2017.

\bibitem[Kim et~al.(2020)Kim, Moon, Rohrbach, Darrell, and Canny]{Kim2020-wp}
Jinkyu Kim, Suhong Moon, Anna Rohrbach, Trevor Darrell, and John Canny.
\newblock Advisable learning for self-driving vehicles by internalizing
  observation-to-action rules.
\newblock In \emph{2020 {IEEE/CVF} Conference on Computer Vision and Pattern
  Recognition ({CVPR})}. IEEE, June 2020.

\bibitem[Koh and Liang(2017)]{Koh2017-xx}
Pang~Wei Koh and Percy Liang.
\newblock Understanding black-box predictions via influence functions.
\newblock In \emph{International conference on machine learning}, pages
  1885--1894. PMLR, 2017.

\bibitem[Koh et~al.(2020)Koh, Nguyen, Tang, Mussmann, Pierson, Kim, and
  Liang]{Koh2020-mp}
Pang~Wei Koh, Thao Nguyen, Yew~Siang Tang, Stephen Mussmann, Emma Pierson, Been
  Kim, and Percy Liang.
\newblock Concept bottleneck models.
\newblock \emph{arXiv preprint arXiv:2007.04612}, July 2020.

\bibitem[Lample and Chaplot(2017)]{Lample2017-em}
Guillaume Lample and Devendra~Singh Chaplot.
\newblock Playing {FPS} games with deep reinforcement learning.
\newblock In \emph{Proceedings of the {Thirty-First} {AAAI} Conference on
  Artificial Intelligence}, AAAI'17, pages 2140--2146, San Francisco,
  California, USA, 2017. AAAI Press.

\bibitem[Lipton(2018)]{Lipton2018-at}
Zachary~C Lipton.
\newblock The mythos of model interpretability: In machine learning, the
  concept of interpretability is both important and slippery.
\newblock \emph{Queueing Syst.}, 16\penalty0 (3):\penalty0 31--57, 2018.

\bibitem[Madumal et~al.(2020)Madumal, Miller, Sonenberg, and
  Vetere]{Madumal2020-ta}
Prashan Madumal, Tim Miller, Liz Sonenberg, and Frank Vetere.
\newblock Explainable reinforcement learning through a causal lens.
\newblock In \emph{Proceedings of the {AAAI} conference on artificial
  intelligence}, volume~34, pages 2493--2500, 2020.

\bibitem[Matuszek et~al.(2013)Matuszek, Herbst, Zettlemoyer, and
  Fox]{Matuszek2013-in}
Cynthia Matuszek, Evan Herbst, Luke Zettlemoyer, and Dieter Fox.
\newblock Learning to parse natural language commands to a robot control
  system.
\newblock In \emph{Experimental robotics}, pages 403--415, 2013.

\bibitem[McGrath et~al.(2021)McGrath, Kapishnikov, Toma{\v s}ev, Pearce,
  Hassabis, Kim, Paquet, and Kramnik]{McGrath2021-qx}
Thomas McGrath, Andrei Kapishnikov, Nenad Toma{\v s}ev, Adam Pearce, Demis
  Hassabis, Been Kim, Ulrich Paquet, and Vladimir Kramnik.
\newblock Acquisition of chess knowledge in {AlphaZero}.
\newblock \emph{arXiv preprint arXiv:2111.09259}, November 2021.

\bibitem[Miller(2019)]{Miller2019-la}
Tim Miller.
\newblock Explanation in artificial intelligence: Insights from the social
  sciences.
\newblock \emph{Artificial intelligence}, 267:\penalty0 1--38, 2019.

\bibitem[Mirowski et~al.(2017)Mirowski, Pascanu, Viola, Soyer, Ballard, Banino,
  Denil, Goroshin, Sifre, Kavukcuoglu, Kumaran, and Hadsell]{Mirowski2017-ay}
Piotr~W Mirowski, Razvan Pascanu, Fabio Viola, Hubert Soyer, Andy Ballard,
  Andrea Banino, Misha Denil, Ross Goroshin, L~Sifre, Koray Kavukcuoglu,
  Dharshan Kumaran, and Raia Hadsell.
\newblock Learning to navigate in complex environments.
\newblock \emph{arXiv preprint arXiv:1611.03673}, 2017.

\bibitem[Mordvintsev et~al.(2015)Mordvintsev, Olah, and
  Tyka]{Mordvintsev2015-bw}
Alexander Mordvintsev, Christopher Olah, and Mike Tyka.
\newblock Inceptionism: Going deeper into neural networks.
\newblock \emph{Google Research Blog}, 2015.

\bibitem[Mott et~al.(2019)Mott, Zoran, Chrzanowski, Wierstra, and
  Jimenez~Rezende]{Mott2019-kg}
Alexander Mott, Daniel Zoran, Mike Chrzanowski, Daan Wierstra, and Danilo
  Jimenez~Rezende.
\newblock Towards interpretable reinforcement learning using attention
  augmented agents.
\newblock \emph{Adv. Neural Inf. Process. Syst.}, 32, 2019.

\bibitem[Olsson et~al.(2022)Olsson, Elhage, Nanda, Joseph, DasSarma, Henighan,
  Mann, Askell, Bai, Chen, Conerly, Drain, Ganguli, Hatfield-Dodds, Hernandez,
  Johnston, Jones, Kernion, Lovitt, Ndousse, Amodei, Brown, Clark, Kaplan,
  McCandlish, and Olah]{Olsson2022-wo}
Catherine Olsson, Nelson Elhage, Neel Nanda, Nicholas Joseph, Nova DasSarma,
  Tom Henighan, Ben Mann, Amanda Askell, Yuntao Bai, Anna Chen, Tom Conerly,
  Dawn Drain, Deep Ganguli, Zac Hatfield-Dodds, Danny Hernandez, Scott
  Johnston, Andy Jones, Jackson Kernion, Liane Lovitt, Kamal Ndousse, Dario
  Amodei, Tom Brown, Jack Clark, Jared Kaplan, Sam McCandlish, and Chris Olah.
\newblock In-context learning and induction heads.
\newblock \emph{Transformer Circuits Thread}, March 2022.

\bibitem[Park et~al.(2018)Park, Hendricks, Akata, Rohrbach, Schiele, Darrell,
  and Rohrbach]{Park2018-ty}
Dong~Huk Park, Lisa~Anne Hendricks, Zeynep Akata, Anna Rohrbach, Bernt Schiele,
  Trevor Darrell, and Marcus Rohrbach.
\newblock Multimodal explanations: Justifying decisions and pointing to the
  evidence.
\newblock In \emph{2018 {IEEE/CVF} Conference on Computer Vision and Pattern
  Recognition}. IEEE, June 2018.

\bibitem[Ravichander et~al.(2020)Ravichander, Belinkov, and
  Hovy]{Ravichander2020-rc}
Abhilasha Ravichander, Yonatan Belinkov, and Eduard Hovy.
\newblock Probing the probing paradigm: Does probing accuracy entail task
  relevance?
\newblock \emph{arXiv preprint arXiv:2005.00719}, 2020.

\bibitem[Ribeiro et~al.(2018)Ribeiro, Singh, and Guestrin]{Ribeiro2018-xx}
Marco~Tulio Ribeiro, Sameer Singh, and Carlos Guestrin.
\newblock Anchors: High-precision model-agnostic explanations.
\newblock In \emph{Proceedings of the AAAI conference on artificial
  intelligence}, volume~32, 2018.

\bibitem[Roh et~al.(2020)Roh, Paxton, Pronobis, Farhadi, and Fox]{Roh2020-zd}
Junha Roh, Chris Paxton, Andrzej Pronobis, Ali Farhadi, and Dieter Fox.
\newblock Conditional driving from natural language instructions.
\newblock In \emph{Conference on Robot Learning}, pages 540--551, 2020.

\bibitem[Rudin(2019)]{Rudin2019-ua}
Cynthia Rudin.
\newblock Stop explaining black box machine learning models for high stakes
  decisions and use interpretable models instead.
\newblock \emph{Nature Machine Intelligence}, 1\penalty0 (5):\penalty0
  206--215, 2019.

\bibitem[Shu et~al.(2017)Shu, Xiong, and Socher]{Shu2017-cb}
Tianmin Shu, Caiming Xiong, and Richard Socher.
\newblock Hierarchical and interpretable skill acquisition in multi-task
  reinforcement learning.
\newblock \emph{arXiv preprint arXiv:1712.07294}, December 2017.

\bibitem[Spirtes et~al.(2000)Spirtes, Glymour, Scheines, and
  Heckerman]{Spirtes2000-xx}
Peter Spirtes, Clark~N Glymour, Richard Scheines, and David Heckerman.
\newblock \emph{Causation, prediction, and search}.
\newblock MIT press, 2000.

\bibitem[Tenney et~al.(2020)Tenney, Wexler, Bastings, Bolukbasi, Coenen,
  Gehrmann, Jiang, Pushkarna, Radebaugh, Reif, and Yuan]{Tenney2020-zb}
Ian Tenney, James Wexler, Jasmijn Bastings, Tolga Bolukbasi, Andy Coenen,
  Sebastian Gehrmann, Ellen Jiang, Mahima Pushkarna, Carey Radebaugh, Emily
  Reif, and Ann Yuan.
\newblock The language interpretability tool: Extensible, interactive
  visualizations and analysis for {NLP} models.
\newblock \emph{arXiv preprint arXiv:2008.05122}, August 2020.

\bibitem[Topin and Veloso(2019)]{Topin2019-nz}
Nicholay Topin and Manuela Veloso.
\newblock Generation of policy-level explanations for reinforcement learning.
\newblock In \emph{Proceedings of the {AAAI} Conference on Artificial
  Intelligence}, volume~33, pages 2514--2521, 2019.

\bibitem[Vygotsky(2012)]{Vygotsky2012-se}
L~S Vygotsky.
\newblock \emph{Thought and Language}.
\newblock The MIT Press. MIT Press, London, England, 2012.

\bibitem[Ward et~al.(2020)Ward, Bolt, Hemmings, Carter, Sanchez, Barreira,
  Noury, Anderson, Lemmon, Coe, Trochim, Handley, and Bolton]{Ward2020-ls}
Tom Ward, Andrew Bolt, Nik Hemmings, Simon Carter, Manuel Sanchez, Ricardo
  Barreira, Seb Noury, Keith Anderson, Jay Lemmon, Jonathan Coe, Piotr Trochim,
  Tom Handley, and Adrian Bolton.
\newblock Using unity to help solve intelligence.
\newblock \emph{arXiv preprint arXiv:2011.09294}, November 2020.

\bibitem[Wiegreffe and Pinter(2019)]{Wiegreffe2019-oa}
Sarah Wiegreffe and Yuval Pinter.
\newblock Attention is not not explanation.
\newblock \emph{arXiv preprint arXiv:1908.04626}, August 2019.

\bibitem[Yau et~al.(2020)Yau, Russell, and Hadfield]{Yau2020-bz}
Herman Yau, Chris Russell, and Simon Hadfield.
\newblock What did you think would happen? explaining agent behaviour through
  intended outcomes.
\newblock \emph{Adv. Neural Inf. Process. Syst.}, 33:\penalty0 18375--18386,
  2020.

\bibitem[Yeh et~al.(2018)Yeh, Kim, Yen, and Ravikumar]{Yeh2018-xx}
Chih-Kuan Yeh, Joon Kim, Ian En-Hsu Yen, and Pradeep~K Ravikumar.
\newblock Representer point selection for explaining deep neural networks.
\newblock \emph{Advances in neural information processing systems}, 31, 2018.

\bibitem[Zambaldi et~al.(2018)Zambaldi, Raposo, Santoro, Bapst, Li, Babuschkin,
  Tuyls, Reichert, Lillicrap, Lockhart, and {Others}]{Zambaldi2018-rl}
Vinicius Zambaldi, David Raposo, Adam Santoro, Victor Bapst, Yujia Li, Igor
  Babuschkin, Karl Tuyls, David Reichert, Timothy Lillicrap, Edward Lockhart,
  and {Others}.
\newblock Deep reinforcement learning with relational inductive biases.
\newblock In \emph{International conference on learning representations}, 2018.

\bibitem[Zeiler and Fergus(2014)]{Zeiler2014-wy}
Matthew~D Zeiler and Rob Fergus.
\newblock Visualizing and understanding convolutional networks.
\newblock In \emph{European conference on computer vision}, pages 818--833,
  2014.

\bibitem[Zhou et~al.(2018)Zhou, Sun, Bau, and Torralba]{Zhou2018-gs}
Bolei Zhou, Yiyou Sun, David Bau, and Antonio Torralba.
\newblock Interpretable basis decomposition for visual explanation.
\newblock In \emph{Proceedings of the European Conference on Computer Vision
  ({ECCV})}, September 2018.

\end{thebibliography}

\appendix

\newpage

\section{Agent architecture and hyperparameters}
\label{appendix-architecture}

We note that our emphasis in this work was not on finding the overall best performing networks, so we did not extensively tune network and learning hyperparameters.

We trained agents using a distributed RL setup, with 4096 parallel actors. We trained the one-hot form using a 4x4 TPUv2, and the language form using a 4x4 TPUv3. Training runs took approximately 12-48 hours to reach maximum episode return ($\sim 200$/episode), typically after 50-200k learner steps.

\begin{table}[htp]
\centering
\begin{tabular}{@{}cclll@{}}
\toprule
\multirow{15}{*}{\begin{tabular}[c]{@{}c@{}}State update\\ $f_\theta$\end{tabular}} &
  \multirow{8}{*}{\begin{tabular}[c]{@{}c@{}}Image\\ encoder\\ $e^{i}_{\theta}$\end{tabular}} &
  \multicolumn{2}{l}{Input resolution} &
  $(160, 192, 3)$ \\ \cmidrule(l){3-5} 
 &
   &
  \multicolumn{1}{l|}{\multirow{7}{*}{ResNet}} &
  number of blocks &
  $3$ \\
 &
   &
  \multicolumn{1}{l|}{} &
  channels per block &
  $(16, 32, 32)$ \\
 &
   &
  \multicolumn{1}{l|}{} &
  conv layers per block &
  $(2, 2, 2)$ \\
 &
   &
  \multicolumn{1}{l|}{} &
  conv filter size &
  $3$ \\
 &
   &
  \multicolumn{1}{l|}{} &
  nonlinearity &
  ReLU \\
 &
   &
  \multicolumn{1}{l|}{} &
  max-pool filter size &
  $3$ \\
 &
   &
  \multicolumn{1}{l|}{} &
  max-pool strides &
  $2$ \\ \cmidrule(l){2-5} 
 &
  \multirow{5}{*}{\begin{tabular}[c]{@{}c@{}}String\\ encoder\\ $e^{s}_{\theta}$\end{tabular}} &
  \multicolumn{1}{l|}{\multirow{3}{*}{Tokenizer}} &
  tokenizer name &
  subword \\
 &
   &
  \multicolumn{1}{l|}{} &
  vocabulary size &
  $8000$ \\
 &
   &
  \multicolumn{1}{l|}{} &
  max token length &
  $19$ (right-padded) \\ \cmidrule(l){3-5} 
 &
   &
  \multicolumn{1}{l|}{\begin{tabular}[c]{@{}l@{}}Linear\\ embedding\end{tabular}} &
  embeddings per token &
  $16$ \\ \cmidrule(l){3-5} 
 &
   &
  \multicolumn{1}{l|}{LSTM} &
  hidden units &
  $256$ \\ \cmidrule(l){2-5} 
 &
  \multirow{2}{*}{\begin{tabular}[c]{@{}c@{}}Memory\\ core\end{tabular}} &
  \multicolumn{2}{l}{Input structure} &
  $[e^{i}_{\theta}(i_t),  e^{s}_{\theta}(s_t), a_{t-1}, r_{t-1}]$ \\ \cmidrule(l){3-5} 
 &
   &
  \multicolumn{1}{l|}{LSTM} &
  hidden units &
  $512$ \\ \midrule
\multicolumn{2}{c}{\multirow{3}{*}{\begin{tabular}[c]{@{}c@{}}Policy head\\ $h_\theta$\end{tabular}}} &
  \multicolumn{1}{l|}{\multirow{2}{*}{Policy MLP}} &
  hidden units &
  $200$ \\
\multicolumn{2}{c}{} &
  \multicolumn{1}{l|}{} &
  action space &
  $\in [-1, 1]^4$ \\ \cmidrule(l){3-5} 
\multicolumn{2}{c}{} &
  \multicolumn{1}{l|}{Value MLP} &
  hidden units &
  $200$ \\ \midrule
\multicolumn{2}{c}{\begin{tabular}[c]{@{}c@{}}CST head\\ $g_\theta$\end{tabular}} &
  \multicolumn{1}{l|}{MLP} &
  hidden units &
  \begin{tabular}[c]{@{}l@{}}$32$ (one-hot)\\ $512$ (language)\end{tabular} \\ \bottomrule
\end{tabular}
\vspace{0.2cm}
\caption{Agent architecture.}
\par
\vspace{1.0cm}
\begin{tabular}{@{}c|ll@{}}
\toprule
\multirow{4}{*}{\begin{tabular}[c]{@{}c@{}}V-Trace\\ Loss\end{tabular}}   & baseline cost                                                  & $1.0$       \\
                          & entropy cost      & $0.001$      \\
                          & $\gamma$          & $0.95$         \\
                          & max reward        & $1.0$               \\ \midrule
\multirow{4}{*}{\begin{tabular}[c]{@{}c@{}}Adam\\ Optimizer\end{tabular}} & learning rate                                                  & $1e^{-4}$ \\
                          & $\beta_1$         & $0.0$               \\
                          & $\beta_2$         & $0.95$              \\
                                                                          & \begin{tabular}[c]{@{}l@{}}clip grad norm\\ above\end{tabular} & $40$      \\ \midrule
\multirow{2}{*}{Schedule} & batch size        & $192$               \\
                          & termination steps & $6e^{7}$ \\ \bottomrule
\end{tabular}
\vspace{0.2cm}
\caption{\label{tab:table-name} Training hyperparameters.}
\end{table}

\newpage

\end{document}